\documentclass{article}
\usepackage{spconf,amsmath,graphicx,url}
\usepackage{multirow}
\usepackage{makecell}
\usepackage{color}

\title{Discriminative analysis of the human cortex using spherical CNNs - \\ a study on Alzheimer's disease diagnosis}

 \name{\normalsize Xinyang Feng * $^{1}$ \quad Jie Yang * $^{1}$ \thanks{* denotes equal contribution.} \quad
 	 Andrew F. Laine $^{1}$ \quad Elsa D. Angelini $^{1,2}$}
 \address{\normalsize $^{1}$ Department of Biomedical Engineering, Columbia University, NY, USA \\
 	 \normalsize $^{2}$ NIHR Imperial BRC, ITMAT Data Science Group, Imperial College, London, UK}

\begin{document}
%
\maketitle
\begin{abstract}
In neuroimaging studies, the human cortex is commonly modeled as a sphere to preserve the topological structure of the cortical surface.
Cortical neuroimaging measures hence can be modeled in spherical representation.
In this work, we explore analyzing the human cortex using spherical CNNs
in an Alzheimer's disease (AD) classification task using cortical morphometric measures derived from structural MRI.
Our results show superior performance in classifying AD versus cognitively normal and in predicting MCI progression within two years, using structural MRI information only.
This work demonstrates for the first time the potential of the spherical CNNs framework in the discriminative analysis of the human cortex
and could be extended to other modalities and other neurological diseases.
\end{abstract}
\section{Introduction}
\label{sec:intro}
Deep learning has witnessed great success in image recognition \cite{he2015delving} using convolutional neural networks (CNNs) and has been widely explored in neuroimaging field \cite{vieira2017dlreview}.
Most previous studies in neuroimaging field either study the extracted features from predefined regions of interest (ROIs) \cite{autism} or feed 3D convolutional neural networks directly with the 3D imaging volume.
The former approach potentially introduces too much prior into the model and limits the input representation.
While the latter approach has the advantage of being agnostic and prior-free, adequate priors from previous neuroimaging studies could be helpful to regularize the input information.

The human cortex is commonly modeled as a 2D manifold sheet-like structure, despite the presence of sulci/gyri folds.
Therefore, 2D CNNs can, in principle, be applied on the cortical sheet after flattening onto a 2D plane \cite{fischl1999cortical}. However, inevitable distortions in the flattening process affect the data representation.
Surface cutting has been proposed to alleviate distortions caused by the intrinsic curvature of the cortical surface,
but this again introduces artificial changes to the topology of the surface \cite{fischl1999cortical}.
Modeling the cortical surface of each hemisphere with a sphere is more accurate and desirable \cite{fischl1999cortical},
and spherical coordinate system is the common practice in neuroimaging field,
as it can preserve the topological structure of the cortical surface.
But 2D CNNs cannot be directly applied on a sphere.

A spherical CNNs framework was recently introduced \cite{s2cnn} and is explored for the first time in this study to analyze the human cortex in a spherical representation.
Spherical CNNs were proposed to model spherical data such as molecular modeling, 3D shape \cite{s2cnn,eccv_scnn} and has shown promising performances.

Alzheimer's disease (AD) is a neurodegenerative disease impacting a large population and is the most common cause of dementia.
Accurate diagnosis of AD and mild cognitive impairment (MCI) is of increasing importance.
Cortical morphometric measures such as cortical thickness derived from T1-weighted structural MRI
have demonstrated to be important biomarkers for the diagnosis of AD, MCI, which are characterized by cortical gray matter atrophy.
In this work, we apply a spherical CNNs based framework on the cortical thickness data derived from structural MRI in Alzheimer's Disease Neuroimaging Initiative (ADNI)\footnote{\url{http://adni.loni.usc.edu/}} cohort,
for the AD versus cognitively normal (CN) classification task, and for MCI conversion prediction within two years.


To the best of our knowledge, this is the first work applying spherical CNNs on human cortex data and demonstrates the potential for diverse studies on discriminative analyses of human cortex neuroimaging data.

\begin{figure*}[!hbtp]
\centering
\includegraphics[width=17.5cm]{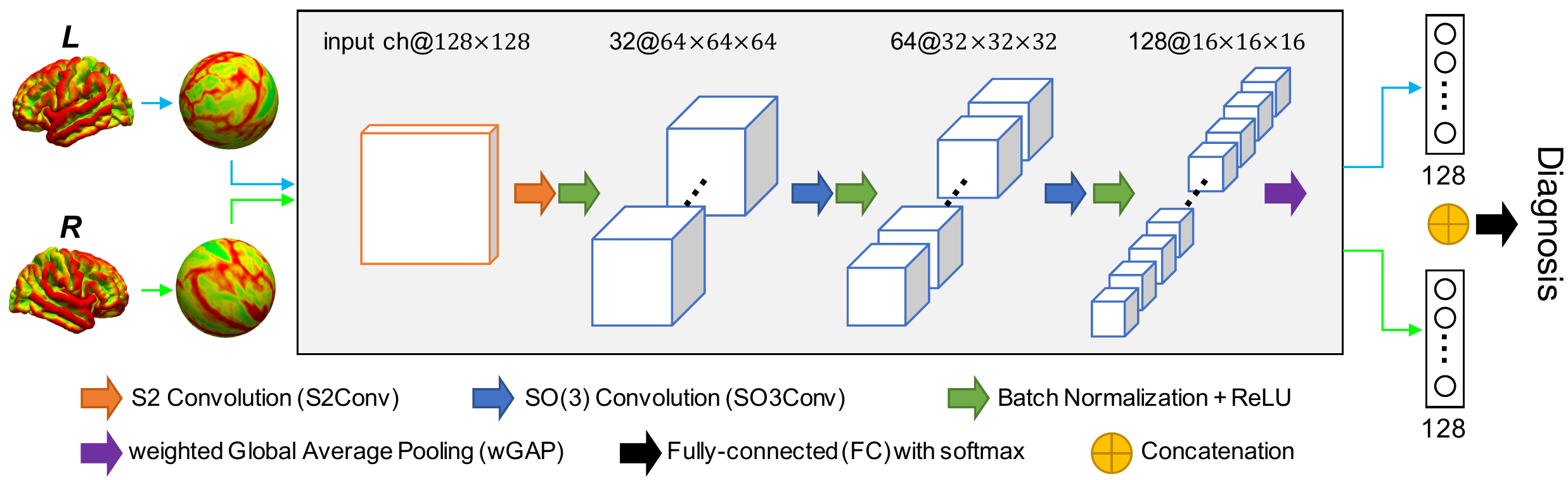}
\vspace*{-3mm}
\caption{\footnotesize{Illustration of the spherical CNNs framework proposed for AD diagnosis based on cortical morphometric data. The basic operation blocks are denoted as arrows and listed under the network structure.}}
\label{Fig:framework}
\end{figure*}

\section{Method}
\label{sec:method}
\subsection{Cortical Modeling}
The cortical surfaces were reconstructed using FreeSurfer \cite{dale1999cortical} and morphed to the spherical representation by minimizing areal and distance distortions.
All the individual cortical surfaces were registered to a spherical atlas in the \textit{fsaverage} space matching cortical folding patterns \cite{fischl1999cortical}.
At each vertex of the atlas cortical surface, multiple measures including thickness, surface area, volume, curvature, sulc, Jacobian determinant (warping to the atlas) can be derived from FreeSurfer.
Sensitivities of different measures vary in different diseases.
And any measure can naturally be regarded as the channels in the data representation.
In this study, we used cortical thickness as it has been previously demonstrated to be highly sensitive for AD diagnosis \cite{tenmethod,eskildsen2013prediction}.


We used a sampling grid with a bandwidth of $64$ to sample the cortical surfaces,
generating a $128\times 128$ matrix for each hemisphere.
For each point in the sampling grid, we queried the closest $10$ vertices in the cortical surface in geodesic distance and used the average measure as the matrix value.



\subsection{Spherical CNNs}
Spherical CNNs are extensions of regular CNNs formulation on the plane to spherical data, migrating the translational equivariance to rotational equivariance.
Hence, specially-designed convolution operations are re-formulated on sphere space S2 and 3D rotation group space SO(3) (SO=`special orthogonal group').
More theoretical underpinnings can be found in \cite{s2cnn}.

Elements in the SO(3) space are represented in the Euler ZYZ data format as:
\vspace{0cm}
\begin{equation}
Z(\alpha)Y(\beta)Z(\gamma)
\end{equation}
\vspace{-0.6cm}

\noindent where $Z(\cdot)$ denotes rotation around the $Z$ axis, $Y(\cdot)$ denotes rotation around the $Y$ axis, $\alpha \in [0,2\pi], \beta \in [0,\pi], \gamma \in [0,2\pi]$ are the rotation angles.

Elements in the S2 space can be similarly represented as:
\vspace{-0.1cm}
\begin{equation}
Z(\alpha)Y(\beta)Z(0)
\end{equation}
\vspace{-0.6cm}

The network architecture is similar to regular CNNs, with spherical convolutional blocks hierarchically layered.
The main parameters include bandwidth $b$, which is similar to the spatial dimension in regular CNNs, and number of channels $c$ at each convolution block.

In this work, we use a simple network structure with three convolutional layers interleaved with 3D batch normalization (BN) and rectifier linear unit (ReLU).
Illustration of the network structure is shown in Fig. \ref{Fig:framework}.
The number of channels doubles and the spatial dimensions reduce by two along the depth.
Specifically, we denote the S2 convolution with bandwidth $b$ and channel $c$ as S2Conv($b$, $c$),
and the SO(3) convolution with bandwidth $b$ and channel $c$ as SO3Conv($b$, $c$).
The fully convolutional part of the network is sequenced as: S2Conv(32, 32) - BN - ReLU - SO3Conv(16, 64) - BN - ReLU - SO3Conv(8, 128) - BN - ReLU.
The three dimensions $\alpha,\beta,\gamma$ of the feature maps at each layer are all $2b$.

Then we apply a weighted global average pooling (wGAP) step,
consisting of integrating over the spatial dimensions of the convolutional feature maps and correcting for the non-uniformity of the grid in the Y axis.

The two hemispheres of human cortex are considered as two sets of spherical data sharing the same diagnosis label.
We therefore share the fully convolutional part of the network between left and right hemispheres.
The integrated features from left and right hemispheres are concatenated and fed into the last fully connected layer with softmax activation function for the final disease classification.

We also compared to regular CNNs on the same sampled input,
with the same architecture using regular 2D convolutions, replacing 3D BN with 2D BN, and doubling the channel dimensions to ensure approximately same number of parameters.
Denoting the convolution operation with $c$ channels as Conv($c$),
the fully convolutional part of the network tested for comparison is:
Conv(64) - BN - ReLU - Conv(128) - BN - ReLU - Conv(256) - BN - ReLU.
The convolution layers have a stride of $2$ and a kernel size of $3$.


For model training, we used the cross-entropy loss and optimized using stochastic gradient descent (SGD) with moment 0.9.
We used a batch size of 8 and ran the algorithm for 200 epochs, with a 0.1 learning rate at the first 100 epochs and 0.01 learning rate for the last 100 epochs.

\subsection{Activation Maps}
Spatial localization of features being used by CNNs can be explored using class activation map \cite{cam}, which has been applied in medical imaging field \cite{nam}.
In this study, we extend the class activation map to spherical CNNs, generating class activation maps on the sphere.
The activation maps in spherical CNNs are defined in SO(3) space.
According to Equation 1 and 2, we selected the activation maps at $\gamma=0$ to explore the activation map patterns and corrected for the non-uniformity of the grid in the Y axis similar to the practice in wGAP.
We performed weighted average of the corrected activation maps using the weights from the fully-connected layer.

\section{Result}
\label{sec:result}
\subsection{Data and Setup}
We used the data from ADNI-1 cohort.
We screened subjects per diagnosis group as follows: CN subjects as having stayed cognitively normal during a follow-up period of at least two years,
MCI stable (MCI-s) subjects as having stayed MCI during a follow-up period of at least two years,
MCI progression (MCI-p) subjects as having converted to AD within two years,
and AD patients.
Subject information for each diagnosis group can be found in Table \ref{Table:adniscan}.

\begin{table}[!hbtp]
\centering
\small{
\caption{\small{Subject Information}}\label{Table:adniscan}
\begin{tabular}{c|c|c|c|c|c}
 \hline
Diagnosis & CN & \makecell{MCI-s} & \makecell{MCI-p} & AD & Total\\
 \hline
N & 151 & 114 & 136 & 188 & 589 \\ \hline
 \makecell{Age\\(std)} & \makecell{75.64 \\(5.25)} & \makecell{74.90\\(7.33)} & \makecell{74.69\\(6.95)} & \makecell{75.18\\(7.50)} & \makecell{75.13\\(6.82)} \\ \hline
 \makecell{Gender\\M/F} & 74/77 & 72/42 & 85/51 & 99/89 & 330/259 \\ \hline
\end{tabular}}
\end{table}

We used the baseline T1-weighted MRI scans acquired using 1.5 T MRI scanners,
pre-processed with the standard Mayo Clinic pipeline\footnote{http://adni.loni.usc.edu/methods/mri-analysis/mri-pre-processing/}
and post-processed by UCSF using FreeSurfer 4.3 \cite{jack2008adni}. 

We performed two binary classification tasks: AD vs. CN and MCI-p vs. MCI-s.
In each classification task, we performed 10-fold cross-validation with the fold split generated from random stratified sampling ensuring similar distribution of diagnosis, age, and gender in each split.

In each experiment, we set out one fold as test set, one fold as validation set, and the rest of the folds as training set.
At each fold, the model with the maximum validation accuracy is selected as the optimal model.
The probability output of all test sets using the optimal models are aggregated together.
We reported the area under curve (AUC) of the receiver operating characteristic (ROC) curve in Fig. \ref{fig:roc},
and also reported the accuracy, sensitivity and specificity.

\subsection{AD vs. CN classification}
\begin{figure}[t]
\center
\includegraphics[width=7.5cm]{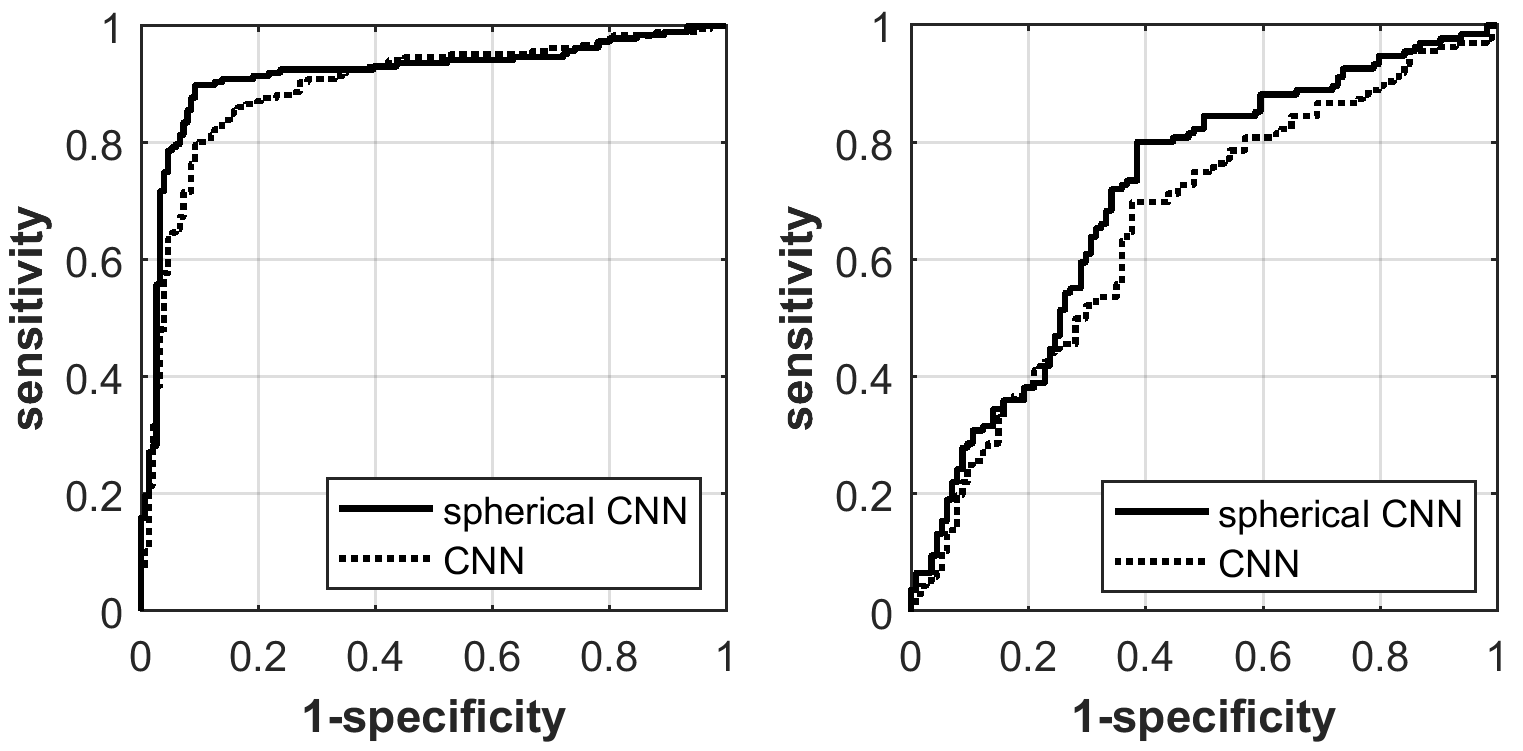}
\vspace*{-3mm}
\caption[ ] {\small{ROC of (left) AD vs. CN classification and (right) MCI-p vs. MCI-s classification.}}
\label{fig:roc}
\end{figure}

The ROC curve for AD vs. CN classification can be found in Fig. \ref{fig:roc} (left). The AUC values for spherical vs. standard CNNs are: 0.915 vs. 0.895. The accuracy (ACC), sensitivity (SEN) and specificity (SPE) values (with 0.5 as threshold) for spherical vs. standard CNNs are: 90.0\% vs. 84.6\%, 89.9\% vs. 84.0\%, 90.1\% vs. 85.4\%. The performance is higher than a previous study also using cortical thickness patterns in ADNI cohort (ACC: 84.5\%, SEN: 79.4\%, SPE: 88.9\%, AUC: 0.905) \cite{eskildsen2013prediction}.

\subsection{MCI progression prediction}
We further test our model on a more challenging MCI progression prediction task using the same network setting. The ROC curve can be found in Fig. \ref{fig:roc} (right). The ROC AUC values for spherical vs. standard CNNs are: 0.707 vs. 0.657. The accuracy, sensitivity and specificity values (with 0.5 as threshold) for spherical vs. standard CNNs are: 71.6\% vs. 66.4\%, 80.2\% vs. 69.9\%, 61.4\% vs. 62.3\%. The performance is higher than a previous study on 2-year MCI progression prediction also using cortical thickness patterns in ADNI cohort (ACC: 66.7\%, SEN: 59.0\%, SPE: 70.2\%, AUC: 0.673) \cite{eskildsen2013prediction}.


\subsection{Exploratory Visualization}
\begin{figure}[t]
\center
\includegraphics[width=7cm]{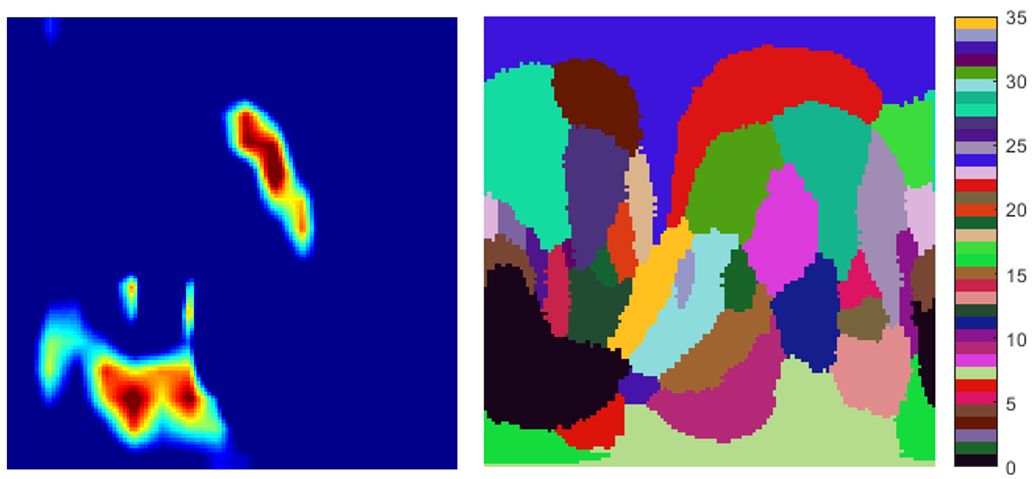}
\vspace*{-3mm}
\caption[ ] {\small{ (Left) Class activation map for AD classification task from the proposed spherical CNN; (Right) Desikan-Killiany atlas in the same space \cite{aparc}.}}
\label{fig:cam}
\end{figure}

A population-average AD class activation map of left hemisphere at $\gamma=0$, generated with the spherical CNNs, is shown in Fig. \ref{fig:cam} together with a reference label map from the Desikan-Killiany atlas \cite{aparc} sampled in the same way as the thickness measures.
The colors and orders of the regions in the reference label map are displayed according to the FreeSurfer color lookup table.
We observed two blobs of AD predictive regions: the lower left blob corresponding to regions around medial temporal lobe,
and the upper blob corresponding to regions in the vicinity of supramarginal gyrus.
Both regions are implicated in AD, according to \cite{regions}.

\section{Discussion}
Despite promising results obtained via our application of spherical CNNs to cortical measures,
there are several limitations and potential future improvements to be considered: 
the input omits subcortical structures, such as the hippocampus, which is one of the brain structures affected by AD and a sensitive biomarker for AD diagnosis \cite{tenmethod,adhc}, 
In future work, the hippocampus can be modeled in the same way as the general 3D structures \cite{s2cnn,eccv_scnn}, and incorporated into the classification model.
And we can use multi-channel input including other measures such as volume to have multi-faceted characterization of the cortex.

We shared the fully convolutional part between left and right hemispheres, while we can also use two different sets of parameters, which however doubles the number of parameters for the network. Left and right hemispheres could also be registered into the same space and concatenated as two channels. By doing so, the asymmetry in the input information could be embedded and utilized by the CNNs for the diagnosis or prediction tasks. 

Since the spherical CNNs formulation is still new to the field, there are still variant architectures to test, such as \cite{eccv_scnn}.
A more thorough exploration of parameters (bandwidth, channel), architectures, and properties (fully-convolutional property) is still necessary to fully exploit its potential.

\vspace{-0.2cm}

\section{Conclusion}
In this study, we demonstrate for the first time that the newly introduced spherical CNNs formulation can be an effective deep learning framework for modeling human cortex and performing AD diagnosis task using MRI-based cortical measures. Our results on the ADNI cohort show state-of-the-art classification performance using structural MRI information only.
The spherical CNNs formulation has the potential to be applied to further structural MRI studies, on other neurological diseases, and other modalities such as fMRI and PET, as long as the measures can be projected onto the cortical sphere.

\vspace{0.2cm}

\noindent\footnotesize{\textbf{Acknowledgments:} Thanks for funding from NIH/NHLBI R01-HL121270. Data used in preparation of this article were obtained from the Alzheimer’s Disease Neuroimaging Initiative (ADNI) database (adni.loni.usc.edu). As such, the investigators within the ADNI contributed to the design and implementation of ADNI and/or provided data but did not participate in analysis or writing of this report. A complete listing of ADNI investigators can be found at: \url{http://adni.loni.usc.edu/wp-content/uploads/how_to_apply/ADNI_Acknowledgement_List.pdf}}


\begin{thebibliography}{10}

\bibitem{he2015delving}
Kaiming He, Xiangyu Zhang, Shaoqing Ren, and Jian Sun,
\newblock ``Delving deep into rectifiers: Surpassing human-level performance on
  imagenet classification,''
\newblock in {\em IEEE International Conference on Computer Vision (ICCV)},
  2015, pp. 1026--1034.

\bibitem{vieira2017dlreview}
Sandra Vieira, Walter H.~L. Pinaya, and Andrea Mechelli,
\newblock ``Using deep learning to investigate the neuroimaging correlates of
  psychiatric and neurological disorders: methods and applications,''
\newblock {\em Neuroscience \& Biobehavioral Reviews}, vol. 74, pp. 58--75,
  2017.

\bibitem{autism}
Heather~Cody Hazlett, Hongbin Gu, Brent~C Munsell, Sun~Hyung Kim, Martin
  Styner, Jason~J Wolff, Jed~T Elison, Meghan~R Swanson, Hongtu Zhu, Kelly~N
  Botteron, et~al.,
\newblock ``Early brain development in infants at high risk for autism spectrum
  disorder,''
\newblock {\em Nature}, vol. 542, no. 7641, pp. 348, 2017.

\bibitem{fischl1999cortical}
Bruce Fischl, Martin~I Sereno, and Anders~M Dale,
\newblock ``Cortical surface-based analysis: {II}: inflation, flattening, and a
  surface-based coordinate system,''
\newblock {\em NeuroImage}, vol. 9, no. 2, pp. 195--207, 1999.

\bibitem{s2cnn}
Taco~S Cohen, Mario Geiger, Jonas Köhler, and Max Welling,
\newblock ``Spherical {CNNs},''
\newblock in {\em International Conference on Learning Representations (ICLR)},
  2018.

\bibitem{eccv_scnn}
Carlos Esteves, Christine Allen-Blanchette, Ameesh Makadia, and Kostas
  Daniilidis,
\newblock ``Learning {SO(3)} equivariant representations with spherical
  {CNNs},''
\newblock in {\em The European Conference on Computer Vision (ECCV)}, September
  2018.

\bibitem{dale1999cortical}
Anders~M Dale, Bruce Fischl, and Martin~I Sereno,
\newblock ``Cortical surface-based analysis: {I}. segmentation and surface
  reconstruction,''
\newblock {\em NeuroImage}, vol. 9, no. 2, pp. 179--194, 1999.

\bibitem{tenmethod}
R{\'e}mi Cuingnet, Emilie Gerardin, J{\'e}r{\^o}me Tessieras, Guillaume Auzias,
  St{\'e}phane Leh{\'e}ricy, Marie-Odile Habert, Marie Chupin, Habib Benali,
  et~al.,
\newblock ``Automatic classification of patients with {A}lzheimer's disease
  from structural {MRI}: a comparison of ten methods using the {ADNI}
  database,''
\newblock {\em NeuroImage}, vol. 56, no. 2, pp. 766--781, 2011.

\bibitem{eskildsen2013prediction}
Simon~F Eskildsen, Pierrick Coup{\'e}, Daniel Garc{\'\i}a-Lorenzo, Vladimir
  Fonov, Jens~C Pruessner, D~Louis Collins, Alzheimer's Disease~Neuroimaging
  Initiative, et~al.,
\newblock ``Prediction of {A}lzheimer's disease in subjects with mild cognitive
  impairment from the {ADNI} cohort using patterns of cortical thinning,''
\newblock {\em NeuroImage}, vol. 65, pp. 511--521, 2013.

\bibitem{cam}
Bolei Zhou, Aditya Khosla, Agata Lapedriza, Aude Oliva, and Antonio Torralba,
\newblock ``Learning deep features for discriminative localization,''
\newblock in {\em IEEE Conference on Computer Vision and Pattern Recognition
  (CVPR)}, 2016.

\bibitem{nam}
Xinyang Feng, Jie Yang, Andrew~F. Laine, and Elsa~D. Angelini,
\newblock ``Discriminative localization in {CNNs} for weakly-supervised
  segmentation of pulmonary nodules,''
\newblock 2017, Medical Image Computing and Computer-Assisted Intervention
  (MICCAI), pp. 568--576.

\bibitem{jack2008adni}
Clifford~R Jack, Matt~A Bernstein, Nick~C Fox, et~al.,
\newblock ``The {A}lzheimer's disease neuroimaging initiative ({ADNI}): {MRI}
  methods,''
\newblock {\em Journal Magn Reson Imaging}, vol. 27, no. 4, pp. 685--691, 2008.

\bibitem{aparc}
Rahul~S Desikan, Florent S{\'e}gonne, Bruce Fischl, Brian~T Quinn, Bradford~C
  Dickerson, Deborah Blacker, Randy~L Buckner, Anders~M Dale, R~Paul Maguire,
  et~al.,
\newblock ``An automated labeling system for subdividing the human cerebral
  cortex on {MRI} scans into gyral based regions of interest,''
\newblock {\em NeuroImage}, vol. 31, no. 3, pp. 968--980, 2006.

\bibitem{regions}
Rahul~S Desikan, Howard~J Cabral, Christopher~P Hess, William~P Dillon,
  Christine~M Glastonbury, Michael~W Weiner, Nicholas~J Schmansky, Douglas~N
  Greve, David~H Salat, et~al.,
\newblock ``Automated {MRI} measures identify individuals with mild cognitive
  impairment and {A}lzheimer's disease,''
\newblock {\em Brain}, vol. 132, no. 8, pp. 2048--2057, 2009.

\bibitem{adhc}
Xinyang Feng, Jie Yang, Andrew~F Laine, and Elsa~D Angelini,
\newblock ``Alzheimer's disease diagnosis based on anatomically stratified
  texture analysis of the hippocampus in structural {MRI},''
\newblock in {\em International Symposium on Biomedical Imaging (ISBI)}. IEEE,
  2018, pp. 1546--1549.

\end{thebibliography}

\end{document}